\pdfoutput=1
\documentclass[11pt]{article}

\PassOptionsToPackage{hyperfootnotes=false}{hyperref}
\usepackage[final]{acl}

\usepackage{times}
\usepackage{latexsym}
\usepackage[T1]{fontenc}
\usepackage[utf8]{inputenc}
\usepackage{microtype}
\usepackage{inconsolata}
\usepackage{graphicx}
\usepackage{booktabs}
\usepackage{amsmath}
\usepackage{url}
\usepackage{xspace}
\usepackage{tabularx}

\newcommand{\system}{\textsc{PROTEA}\xspace}

\title{\system: Offline Evaluation and Iterative Refinement\\
for Multi-Agent LLM Workflows}

\author{
  Kazuki Kawamura, Satoshi Waki, Kei Tateno \\
  Sony Group Corporation \\
  \texttt{\{kazuki.kawamura,satoshi.waki,kei.tateno\}@sony.com}
}

\begin{document}
\maketitle
\begingroup
\renewcommand{\thefootnote}{}
\footnotetext{To appear in the Proceedings of the 64th Annual Meeting of the Association for Computational Linguistics (ACL 2026) System Demonstrations.}
\endgroup

\begin{abstract}
  Multi-agent LLM workflows---systems composed of multiple role-specific LLM calls---often outperform single-prompt baselines, but they remain difficult to debug and refine.
  Failures can originate from subtle errors in intermediate outputs that propagate to downstream nodes, requiring developers to inspect long traces and infer which agent to modify.
  We present \system, a unified interface for offline, test-driven improvement of multi-agent workflows.
  \system executes a workflow, scores intermediate node outputs with configurable rubrics, and overlays per-node states and rationales on the workflow graph to localize likely bottlenecks.
  To support complex systems where final-answer references are the primary supervision, \system performs backward node evaluation: it generates candidate node-level expectations from final-answer references and graph context, then compares them with observed node outputs.
  For selected nodes, \system presents targeted prompt revisions as editable before/after comparisons, then automatically reruns and re-evaluates the workflow to show output changes and score trajectories within the same interface.
  In two production-adjacent workflows, \system improved document-inspection accuracy from 64.3\% to 83.9\% and recommendation Hit@5 from 0.30 to 0.38.
  In a formative study with six experienced LLM developers, participants valued graph-level localization, per-node rationales, and editable before/after prompt revisions.
\end{abstract}

\section{Introduction}

LLM applications are increasingly built as agentic and multi-agent systems that compose multiple role-specific LLM calls into a workflow graph.
By splitting responsibilities across nodes (e.g., intent analysis, retrieval, planning, ranking, and response generation), such graph-based workflows can be more controllable and produce higher-quality outputs than single-prompt approaches.
This design pattern is now widely supported by orchestration frameworks (e.g., AutoGen, LangGraph) and decomposition-based prompting paradigms~\cite{wu2024autogen,langgraph,yao2022react}.

However, this decomposition makes refinement a systems-level debugging task.
When the final answer needs improvement, a small omission, misunderstanding, or formatting drift in an upstream artifact can propagate to downstream nodes.
Two questions dominate day-to-day iteration: which node is a useful starting point for inspection, and what minimal prompt revision would improve the workflow while preserving desired behavior.

\begin{figure}[t]
  \centering
  \includegraphics[width=\columnwidth]{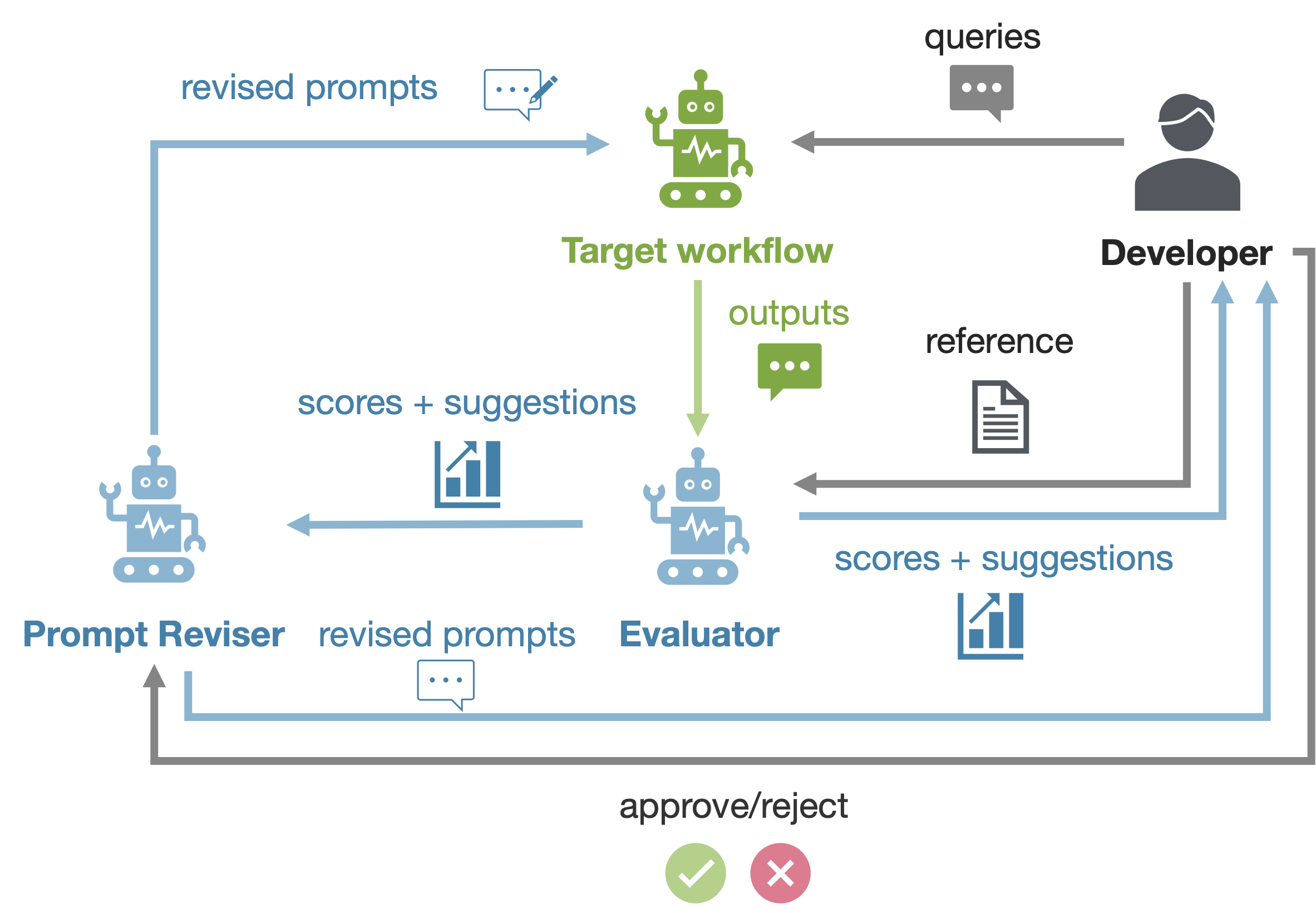}
  \caption{Overview of \system: an interactive, evaluation-driven framework for developer-steered, AI-assisted refinement of multi-agent LLM workflows. Developers identify bottlenecks, inspect evidence and prompt revisions, edit or approve changes, and compare behavior within one loop.}
  \label{fig:overview}
\end{figure}

A growing body of work provides evaluation resources for LLMs and agents, but these typically emphasize end-to-end success rather than graph-aware diagnosis.
Agent benchmarks for tool use and multi-step behavior (e.g., AgentBench and ToolLLM/ToolBench) report task-level outcomes and aggregated statistics~\cite{liu2023agentbench,qin2023toolbench}.
Similarly, practical evaluation frameworks and libraries---including OpenAI Evals, lm-evaluation-harness, OpenEvals, DeepEval, and promptfoo---support offline testing and regression suites for LLM applications~\cite{openai2023evals,lmevalharness,openevals,deepeval,promptfoo}.
For specific architectures such as RAG, frameworks like RAGAs and ARES evaluate components such as retrieval relevance and answer faithfulness~\cite{es2024ragas,saadfalcon2024ares}.
These tools are useful for measurement and comparison, yet they often evaluate the application as an end-to-end system: even when traces are available, the developer must manually determine where in a workflow the error was introduced and how to revise a particular node prompt.

Another line of tools focuses on the development and operation of LLM applications by providing tracing, datasets, evaluations, and prompt management in the interface.
For example, LangSmith aims to support tracing and evaluation alongside prompt testing and deployment management, while platforms such as Arize Phoenix, Weave, Langfuse, and PromptLayer provide complementary combinations of observability, evaluation, and prompt/version management~\cite{langsmith,phoenix,weave,langfuse,promptlayer}.
These platforms substantially improve visibility, but the iteration-and-comparison loop for multi-agent graphs still requires substantial manual effort: developers translate evaluation evidence into a candidate fix, edit prompts (often in a separate view or in code), rerun the workflow, and then reconstruct what changed.
Moreover, in realistic product settings, teams often have labels only for the terminal output; defining reference outputs for every intermediate node is costly, and such references are rarely maintained.

Automatic prompt optimization and self-improvement methods address a different part of the problem.
DSPy compiles modular LM programs and can optimize prompts/modules against an end metric; OPRO and APE use LLMs to search for improved prompts; Self-Refine and Reflexion iteratively improve generations via self-feedback and reflection~\cite{khattab2023dspy,yang2023opro,zhou2023ape,madaan2023selfrefine,shinn2023reflexion}.
These approaches are complementary to \system: they optimize program behavior against an objective, whereas \system focuses on the developer workflow for a concrete multi-agent graph by surfacing candidate bottleneck nodes and proposing localized, inspectable prompt revisions that a developer can adopt, edit, or revert in context.

We present \system, a unified interface for offline evaluation and iterative refinement of multi-agent LLM workflows that is explicitly designed for workflow designers, prompt engineers, and planners.
\system is built around three ideas.
First, it surfaces node-level evidence rather than only terminal scores: intermediate node outputs are evaluated with configurable rubrics using LLM-as-a-judge techniques~\cite{liu2023geval,zheng2023mtbench,gu2024llmjudge_survey}, and pass/warn/fail states plus rationales are overlaid directly on the workflow graph.
Second, \system reduces labeling effort via backward node evaluation: from final-answer references, it generates node-level expectations using the desired final output and downstream requirements, evaluates nodes in reverse topological order, and highlights upstream outputs as candidate bottlenecks for developer inspection.
Third, \system provides localized, adoptable fixes: for a developer-selected node, it proposes targeted prompt revisions in editable before/after comparisons grounded in the evaluator rationale and node rubric.
After the developer adopts (or edits) the prompt revision, \system automatically reruns and re-evaluates the workflow and shows before/after comparisons and score histories in the same interface, enabling rapid iteration without tool switching.
We evaluate \system through two production-adjacent workflow studies, an automatic iteration stress test, and a formative user study with six experienced LLM developers.
The workflow studies improved document-inspection accuracy from 64.3\% to 83.9\% and recommendation Hit@5 from 0.30 to 0.38; the user study suggests that developers value graph-level localization, node-level rationales, and editable before/after prompt revisions for offline debugging and refinement.

\begin{figure*}[t]
  \centering
  \includegraphics[width=\textwidth]{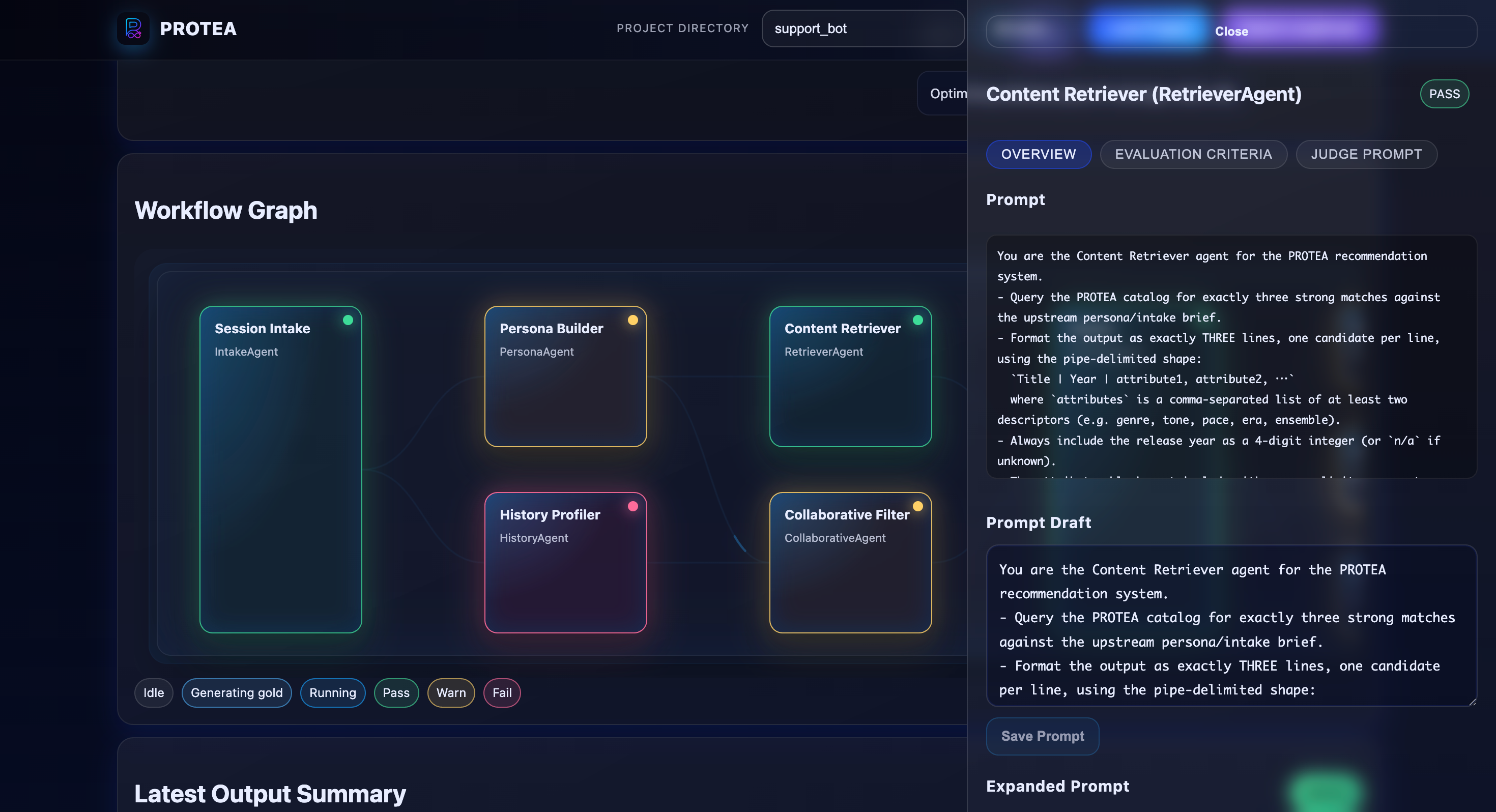}
  \caption[Representative \system interface.]{
Representative \system interface for inspecting and debugging a multi-agent workflow.
The main view shows the workflow graph and node-level evaluation states.
Red/yellow nodes indicate \textsc{fail}/\textsc{warn} states and guide inspection.
The inspection panel shows details for the selected node, including its prompt, evaluation result, suggested revision, and expanded prompt.
Developers can inspect connections, review node behavior, and revise prompts while monitoring workflow-level effects.
}
  \label{fig:protea-representative-screen}
\end{figure*}

\section{System Overview}
\label{sec:overview}

\system supports offline refinement of multi-agent LLM workflows represented as directed acyclic graphs (DAGs).
Each node corresponds to a role-specific LLM call (or a tool-augmented agent) with its own prompt and evaluation criteria, and edges represent intermediate outputs passed between agents.
The system is organized around a fixed-suite iteration loop: run the current workflow, evaluate intermediate node outputs, inspect highlighted nodes, revise the selected node's prompt, and rerun the workflow to check regressions.

\subsection{Workflow setup and saved runs}

A \system project combines (i) a workflow specification (loaded from a saved project or imported from LangGraph), (ii) an offline test suite of queries with optional final-answer references, and (iii) per-node evaluator settings (rubrics, judge prompts, and thresholds).
When human-provided intermediate references are available, they can be attached to the corresponding node.
Otherwise, \system can generate a candidate node-level expectation for each node from the final-answer reference and graph context using the backward mode described in Section~\ref{sec:backward}.
This generated expectation is stored and shown in the interface, so developers can inspect, edit, or override it as needed.

For each run, \system saves node traces (inputs/outputs and metadata), evaluator scores and rationales, any generated node-level expectations, and the current set of node prompts.
These saved records support the interface shown in Figure~\ref{fig:protea-representative-screen} and make iterations comparable on the same test suite.

\subsection{System architecture}

\system consists of a browser-based interface and a backend service that orchestrates (a) workflow execution, (b) automated evaluation using rubric-based LLM judging, and (c) prompt revision assistance.
At a high level (Figure~\ref{fig:overview}), an automated evaluation module produces per-node results and improvement suggestions.
When final-answer references are the available supervision, this module first performs backward node evaluation to generate candidate node-level expectations from the final-answer reference and dependencies between nodes, then compares them to observed node outputs to produce scores and rationales.
An automatic improvement module turns these evaluation results into an editable prompt draft for a selected node.
All prompts, traces, and evaluation outputs are stored with the project so that developers can compare iterations.

\subsection{How a developer uses \system in practice}

\begin{figure*}[t]
  \centering
  \includegraphics[width=\textwidth]{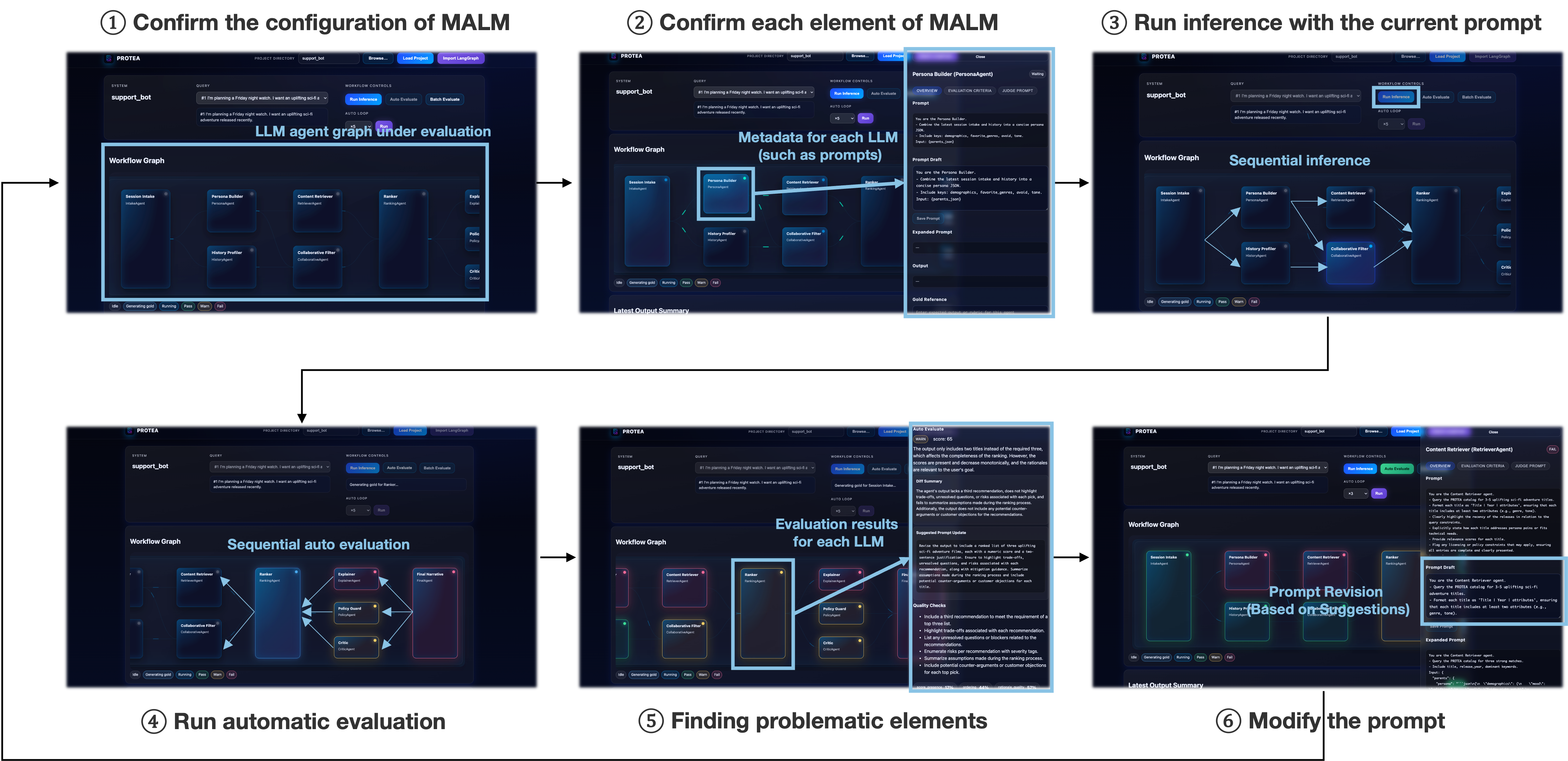}
  \caption{Six-step interface flow for improving a target multi-agent LLM workflow in \system, labeled as MALM (the target multi-agent LLM workflow) in the interface:
    (1) load a project and confirm the agent graph,
    (2) inspect per-node metadata (prompts/criteria/references),
    (3) run inference with the current prompts on a selected test query (or a batch of queries),
    (4) run automatic evaluation (optionally generating node-level expectations when intermediate references are missing),
    (5) inspect per-node results to identify candidate bottleneck nodes, and
  (6) revise the selected node's prompt using the suggested draft.}
  \label{fig:ui_flow}
\end{figure*}

Figure~\ref{fig:ui_flow} illustrates the typical interaction flow.
A developer first loads a project and confirms the target agent graph (Step~1), then clicks nodes to inspect and edit their metadata such as prompts, evaluation criteria, and (when available) references (Step~2).
They run the workflow with the current prompts on a selected test query (Step~3) or a batch of queries, and then trigger \textsc{Auto Evaluate} to score intermediate node outputs (Step~4).
When final-answer references are the available supervision, \textsc{Auto Evaluate} can additionally generate node-level expectations in a backward manner from the final-answer reference and downstream requirements (Section~\ref{sec:backward}); the interface exposes this stage as reference generation and stores the generated expectation for inspection.
The workflow graph is annotated with discrete states (\textsc{pass}/\textsc{warn}/\textsc{fail}), and selecting a highlighted node opens an inspection panel that shows its outputs alongside evaluator rationales and suggested prompt updates (Step~5).
Finally, the developer edits the suggested prompt draft for that node, saves it, and reruns inference and evaluation to compare the new behavior against the same offline suite (Step~6).
Across iterations, the evaluation-history view summarizes score trajectories and enables replaying past runs and restoring previous prompt sets for rollback. Figure~\ref{fig:protea-representative-screen} shows how graph-level states and node-level inspection are combined in the interface. \system also provides an automatic iteration mode (\textsc{Auto Loop} in the interface) to repeat this evaluate$\rightarrow$revise$\rightarrow$re-evaluate cycle for a fixed number of iterations while logging each run.

\section{Key Functionalities}
\label{sec:functionalities}

\system supports multi-agent workflow iteration under realistic constraints---complex dependencies, final-answer-only references, and long trace histories---by coupling three core mechanisms: backward node evaluation for generating node-level expectations, node-level diagnosis support in the workflow interface, and validated prompt refinement with automatic re-evaluation.

\subsection{Node-level Evaluation via Backward Inference}
\label{sec:backward}

Agents in a workflow are coupled through the outputs they pass to downstream nodes.
Developers often use final-answer references as the primary supervision signal, while node-level references are created selectively.
To support this setting, \system uses \textit{Backward Node Evaluation}: starting from final-answer supervision, the system generates a candidate expectation for each node using the requirements of downstream nodes.

Formally, let the workflow be a DAG $G=(V,E)$.
Each node $v \in V$ has an instruction $I_v$ and, when provided, a required output format $\mathcal{S}_v$.
For a given test query, the workflow produces an observed output $y_v$ at each node.
The reference for $v$, denoted $y_v^{\mathrm{ref}}$, is chosen as follows.
For the final node, the final-answer reference $y_{\mathrm{final}}^\star$ is used when available; if it is unavailable, a human-provided node reference $y_v^\star$ is used when present, and otherwise \system generates a fallback candidate reference.
For intermediate nodes, a human-provided node reference $y_v^\star$ is used when available; otherwise, \system generates a candidate reference $\hat{y}_v$ from the final-answer reference and graph context when possible, falling back to the node instruction and required output format.

The candidate reference is produced by a reference-generation step, denoted $\mathcal{M}_{\mathrm{gen}}$.
We use separate symbols for the reference-generation, evaluation, and prompt-revision steps to distinguish their roles in the loop.
For node $v$, the reference-generation step receives a context containing $I_v$, the optional required output format $\mathcal{S}_v$, the node's local position in the graph, the needs of its immediate children, and, when available, the final-answer reference for the current query.
It generates a node output that supports the downstream path toward the desired final answer.

When $v$ has multiple children, their requirements are presented together to produce a consolidated expectation for the node.
In other words, the generated expectation is shaped to be useful to all immediate downstream nodes.
A simple fallback reference derived from the node instruction and required output format is used as a robust default.

Given $y_v$ and $y_v^{\mathrm{ref}}$, an evaluator call $\mathcal{M}_{\mathrm{eval}}$ scores the node output using node-specific criteria.
For each evaluation criterion $d$, the evaluator produces a score $\sigma_d(v) \in [0,1]$, and \system computes an overall weighted score $s(v) \in [0,1]$.
The score is mapped to \textsc{pass}, \textsc{warn}, or \textsc{fail} using configurable thresholds; the default thresholds are $0.8$ for \textsc{pass} and $0.55$ for \textsc{warn}.
The evaluator also returns a short rationale $R_v$ and a suggested direction for improving the node prompt; \system stores these with the reference used for scoring so developers can inspect the basis of each state.

\subsection{Node-level Diagnosis Support}
\label{sec:diagnosis}

In workflows with many nodes, developers need a compact view of where to focus inspection.
\system surfaces node-level evidence on the workflow graph so that the developer can quickly choose a node to inspect.

After evaluation, each node is annotated with its state, score, and evaluator rationale.
The interface highlights nodes requiring attention and sorts the displayed node list by status, with \textsc{fail} before \textsc{warn} before \textsc{pass}; within the same status, nodes with lower scores are shown earlier.
This ordering gives developers a starting point for inspection while keeping the final choice under their control.

When the developer selects a node, \system opens an inspection panel showing the node output, the reference used for evaluation, the evaluator rationale, and the suggested prompt revision.
This design keeps the developer in control while reducing the need to read the full execution trace manually.

\subsection{Prompt Refinement and Re-evaluation}
\label{sec:patch}

To support the end-to-end iteration loop, \system provides prompt refinement assistance for nodes selected by the developer.
Given a node $v$, the prompt-revision step $\mathcal{M}_{\mathrm{opt}}$ proposes a revised instruction $I'_v$.

The prompt-revision step receives the current instruction $I_v$, the evaluator rationale, and a short improvement suggestion derived from the node-level evaluation.
It returns a rewritten prompt and a brief note explaining the change.
The interface presents the result as an editable before/after comparison, so the developer can accept, revise, or discard the proposed prompt revision.

The prompt-revision step is instructed to preserve the variables used to pass the user query and parent-node outputs, keep the expected output format stable, and avoid copying test-specific content into the prompt.
\system also checks proposed prompts for direct copying of test questions and rejects such changes when detected.
Format preservation is ultimately verified by rerunning and re-evaluating the workflow.

After a prompt revision is accepted, \system reruns the full workflow on the offline test suite and evaluates the new outputs.
In automatic iteration mode, \system repeats the evaluate--revise--re-evaluate cycle for a fixed number of iterations.
In this mode, \system retains a proposed prompt revision only when repeated checks show improvement and stable behavior across the fixed suite.

\section{Evaluation}
\label{sec:evaluation}

We evaluate \system through two offline workflow-improvement studies and a formative user study with experienced LLM developers.
The workflow-improvement studies include two internal workflows close to production use, evaluated with a developer-in-the-loop protocol, and an automatic iteration stress test on independently generated workflows.
For the stress test, we include a five-workflow subset with no-rewrite baselines to distinguish the effect of prompt revision from score variation caused by repeated runs.
For the internal workflow studies, we report aggregate outcomes and recurring qualitative patterns while preserving project confidentiality.

\subsection{Production-adjacent workflow evaluations}

We report two internal case studies on workflows close to production use.
Both use a developer-in-the-loop protocol: evaluation and backward expectation generation are automated, while prompt revisions are human-authored and human-approved.

\paragraph{Case study A: enterprise document inspection.}
We applied \system to a high-volume document inspection workflow that must satisfy strict policy and style constraints, check dozens of criteria, and output per-criterion judgments with rationales. Initial prompts and node instructions were written by developers during prior system development (not generated for this study). The workflow is a multi-agent DAG with $N{=}5$ nodes. We evaluated on an internal manually labeled test set of several dozen documents with representative criteria, using item-level correctness (binary match to human labels).
Starting from the initial multi-agent workflow, iterative refinement with \system improved accuracy from 64.3\% to 83.9\%.
In practice, many accepted revisions fell into two categories: making intermediate outputs more explicit and tightening node-specific rubrics. These changes made downstream behavior less ambiguous and evaluator rationales more actionable.

\paragraph{Case study B: conversational recommendation/matching.}
Our second case study is an internal conversational recommendation workflow that clarifies an underspecified request, generates candidate items, ranks them, and produces a final response. As in Case study A, initial prompts and node instructions were developer-written and refinement was developer-in-the-loop; the workflow DAG has $N{=}6$ nodes. Each test case includes a small set of target items, and we evaluate recommendation quality using Hit@5, defined as whether the ranked list contains at least one target item within the top five recommendations.
Across iterative refinement cycles in \system, Hit@5 improved from 0.30 to 0.38.
This workflow uses final-answer references as the primary supervision signal; backward node evaluation and per-node rationales helped trace how constraints propagated through the graph, and localized prompt revision reduced the need to read long traces end-to-end.

\subsection{Automatic iteration stress test}
\label{sec:auto_loop_stress}

The internal workflow evaluations above test \system in developer-in-the-loop settings, where humans inspect and approve prompt revisions.
We also tested \system's automatic iteration mode, \textsc{Auto Loop}, where the system proposes prompt revisions and retains them only when repeated offline checks show improvement.

We collected eleven multi-agent workflows that were not used during \system development.
Each workflow was generated independently by an LLM using only the usage documentation.
All eleven workflows executed end-to-end without execution errors.
For quantitative analysis, we focus on five workflows whose initial node prompts were deliberately minimal, leaving room for automatic refinement.
The five workflows covered HTTP log triage, course scheduling, incident-ticket generation, multi-step word problems, and refuse-or-clarify support responses.

For each workflow, we ran \textsc{Auto Loop} for three iterations on the first query.
As a no-rewrite baseline, we also ran the original workflow three times with prompt rewriting disabled.
This baseline estimates score variation without prompt revisions.

Table~\ref{tab:auto_loop_stress} reports the no-rewrite baseline, the best score observed during automatic iteration, and the corresponding gain.
Overall, four of the five minimal-prompt workflows achieved higher best scores than the no-rewrite baseline in this small stress test, three of those four achieved gains above 0.3 score units, and one workflow did not improve.
The largest improvement occurred on the course-scheduling workflow, where the score increased from 0.186 to 0.800.
The word-problem workflow retained its initial score because its numeric-correctness criterion was effectively exact-match: incorrect final answers received near-zero credit even when intermediate reasoning was partially useful.
As a result, the optimizer received little graded signal indicating which prompt revisions moved the workflow closer to the correct answer.

\begin{table}[t]
\centering
\footnotesize
\setlength{\tabcolsep}{3pt}
\begin{tabular}{lrrr}
\toprule
Workflow & No-rewrite baseline & Best score & Gain \\
\midrule
Log triage & $0.307{\pm}0.029$ & 0.648 & +0.341 \\
Scheduling & $0.186{\pm}0.001$ & 0.800 & +0.614 \\
Incident ticket & $0.333{\pm}0.110$ & 0.840 & +0.507 \\
Refuse/clarify & $0.208{\pm}0.027$ & 0.390 & +0.182 \\
\addlinespace
Word problem & $0.000{\pm}0.000$ & 0.000 & 0.000 \\
\bottomrule
\end{tabular}
\caption{
Automatic iteration stress test. Baseline is mean$\pm$std over three no-rewrite runs; best score is the best three-iteration \textsc{Auto Loop} score.
}
\label{tab:auto_loop_stress}
\end{table}

\subsection{Formative user study with experienced LLM developers}

We ran a formative user study in which participants used \system to debug and refine a conversational recommendation workflow and then provided feedback via a short questionnaire and open-ended comments. Six experienced LLM developers participated. The goal was to identify useful interface features and practical integration needs. We provided a pre-configured workflow and a small offline test suite with examples exhibiting failures such as missing hard constraints or incorrect item ranking; participants diagnosed the likely failing node, applied or edited a suggested prompt revision, and verified changes via before/after comparisons and score trajectories in the history view.

Participants expressed interest in \system for offline iteration and debugging, especially for workflows with many nodes and intermediate outputs. They highlighted (i) graph-level localization for narrowing down candidate nodes, (ii) per-node rationales that make outcomes actionable, and (iii) the ability to adopt prompt revisions through before/after comparison views with automatic reruns.

Participants also identified priorities for adoption: evaluator calibration and generalization checks (alignment of LLM-as-a-judge scores and backward-inferred expectations across data and stochastic runs), integration with existing toolchains (importing graphs/prompts, running regression suites, exporting prompt versions for CI), team operation and reproducibility (clearer provenance and versioning), and the cost/latency of repeated offline evaluation on large suites.
These priorities align with \system's developer-in-the-loop use and motivate calibrated evaluation, richer regression summaries, and smoother pipeline integration.

\section{Conclusion}
We presented \system, a unified interface for offline, test-driven refinement of multi-agent LLM workflows, combining node-level evaluation (including backward reference generation), graph-based diagnosis support, and editable prompt revisions with automatic re-evaluation.
In our case studies, \system improved document inspection accuracy from 64.3\% to 83.9\% and increased Hit@5 from 0.30 to 0.38.
In an automatic iteration stress test, eleven independently generated workflows executed end-to-end without execution errors; in the five-workflow subset with deliberately minimal initial prompts and no-rewrite baselines, four workflows achieved higher best scores than the no-rewrite baseline in this small stress test, and three achieved gains above 0.3 score units.
A formative user study with six experienced LLM developers indicated interest in the approach and highlighted priorities for calibrated evaluation, smoother integration with existing toolchains, and support for richer workflow structures and structured outputs.

\section*{Limitations}

\system is designed primarily for developer-in-the-loop iteration on fixed workflow graphs, with \textsc{Auto Loop} supporting repeated prompt refinement over a fixed regression suite. Rubric-based LLM evaluation benefits from careful evaluator design; judge calibration, multi-judge agreement, and targeted human review loops can further strengthen score stability.

The automatic stress test also showed that near-binary criteria can provide limited feedback to \textsc{Auto Loop}. When a numeric task gives near-zero credit to all incorrect final answers, the optimizer has limited ability to distinguish partially better prompt revisions. This can be mitigated by adding partial-credit criteria for intermediate facts, reasoning setup, constraint satisfaction, or format validity, while keeping exact final correctness as only one component of the rubric.

The current prototype targets DAG-based workflows, with cyclic control flow, supervisor-based coordination, and long-running interactive agents as natural extensions of the execution and visualization model. \system currently focuses on localized prompt revisions within a fixed workflow; future versions can support architectural edits such as adding nodes and rewiring dependencies, together with cost-aware prompt compression.

Larger controlled studies measuring time-to-fix and regression rates, as well as broader stress tests across backbone models and workflow suites, will further characterize the developer workflow and its relation to automated prompt-optimization methods.

\bibliography{custom}

\end{document}